\pdfoutput=1

\documentclass[11pt]{article}

\usepackage[preprint]{acl}

\usepackage{times}
\usepackage{latexsym}
\usepackage{booktabs}
\usepackage[T1]{fontenc}

\usepackage[utf8]{inputenc}

\usepackage{microtype}

\usepackage{inconsolata}

\usepackage{multirow}
\usepackage[ruled,vlined]{algorithm2e}
\usepackage{graphicx}
\usepackage{tabularx}
\usepackage{soul}
\title{Improving Grammatical Error Correction via Contextual Data Augmentation}

\author{Yixuan Wang$^1$, Baoxin Wang$^{1,2}$, Yijun Liu$^1$, Qingfu Zhu$^{1,}$\thanks{Corresponding author.}, Dayong Wu$^2$, Wanxiang Che$^1$ \\
$^1$Research Center for Social Computing and Information Retrieval, Harbin Institute of Technology, China\\
$^2$State Key Laboratory of Cognitive Intelligence, iFLYTEK Research, China \\
\{yixuanwang, yijunliu, qfzhu, car\}@ir.hit.edu.cn \\
\{bxwang2, dywu2\}@iflytek.com\\}

\begin{document}
\maketitle
\begin{abstract}
Nowadays, data augmentation through synthetic data has been widely used in the field of Grammatical Error Correction (GEC) to alleviate the problem of data scarcity.
However, these synthetic data are mainly used in the pre-training phase rather than the data-limited fine-tuning phase due to inconsistent error distribution and noisy labels.
In this paper, we propose a synthetic data construction method based on contextual augmentation, which can ensure an efficient augmentation of the original data with a more consistent error distribution.
Specifically, we combine rule-based substitution with model-based generation, using the generative model to generate a richer context for the extracted error patterns.
Besides, we also propose a relabeling-based data cleaning method to mitigate the effects of noisy labels in synthetic data.
Experiments on CoNLL14 and BEA19-Test show that our proposed augmentation method consistently and substantially outperforms strong baselines and achieves the state-of-the-art level with only a few synthetic data.

\end{abstract}

\section{Introduction}
Grammatical Error Correction (GEC) aims to detect and correct grammatical errors in a text \cite{wang2020comprehensive, bryant2022grammatical}.
It is a challenging task with a wide range of application scenarios, including search engines, writing assistants \cite{omelianchuk2020gector}, and Automatic Speech Recognition (ASR) systems.
Due to the low frequency of grammatical errors in real corpus, obtaining and annotating a certain number of high-quality GEC datasets is usually difficult and costly.
Therefore, the currently available high-quality annotated GEC data is very limited \cite{ye2023mixedit}, making synthetic data an important research direction for the data-starved task.
\begin{figure}[t]
    \centering
    \includegraphics[scale=0.5]{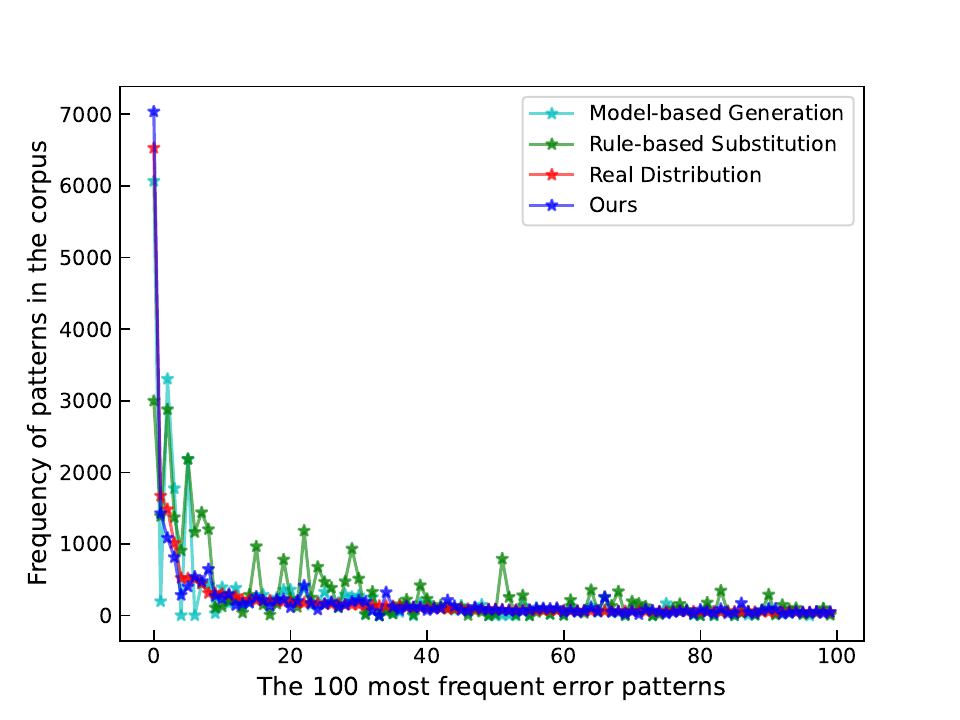}
    \caption{Illustration of the distribution of error patterns in each dataset. The x-axis represents the 100 most frequent error patterns in the annotated dataset W\&I+L, and the y-axis represents the frequency of that error in the corresponding synthetic dataset.}
    \label{fig:distru}
\end{figure}

Nowadays, using synthetic data or data augmentation to improve the performance of GEC models has become a mainstream approach \cite{madnani2012exploring, grundkiewicz2014wiked, grundkiewicz2019neural}.
Common construction methods can be categorized into rule-based substitution \cite{awasthi2019parallel,choe2019neural} and model-based generation methods \cite{xie2018noising,lichtarge2019corpora,zhou2019improving,fang2023transgec}.
However, the synthetic data constructed by the above methods are mainly used in the pre-training phase to initialize a better GEC model.
The data augmentation methods used for the data-limited fine-tuning phase are of great research value.

\begin{table*}[t]
    \centering
    \scalebox{0.9}{
    \begin{tabular}{c|c|c}
    \hline
    \multicolumn{2}{c|}{Wrong Sentence}  &  Public transport enables our body to \textcolor{red}{move one} place to another.\\
    \multicolumn{2}{c|}{Correct Sentence} &  Public transport enables our body to \textcolor{green!70!black}{move from one} place to another.\\
    \hline
    \multirow{3}{*}{1-gram Aug} & Pattern & \textcolor{red}{$\emptyset$} $\rightarrow$ \textcolor{green!70!black}{from}\\
    & Source & They are coming \textcolor{red}{$\emptyset$} the city center.\\
    & Target & They are coming \textcolor{green!70!black}{from} the city center.\\
    \hline
    \multirow{3}{*}{3-gram Aug} & Pattern & \textcolor{red}{move one} $\rightarrow$ \textcolor{green!70!black}{move from one}\\
    & Source & They \textcolor{red}{move one} place to another.\\
    & Target & They \textcolor{green!70!black}{move from one} place to another.\\
    \hline
    \multirow{3}{*}{5-gram Aug} & Pattern & \textcolor{red}{to move one place} $\rightarrow$ \textcolor{green!70!black}{to move from one place}\\
    & Source & They will have \textcolor{red}{to move one place} to another in order to find the treasure.\\
    & Target & They will have \textcolor{green!70!black}{to move from one place} to another in order to find the treasure.\\
    \hline
    \end{tabular}
    }
    \caption{An example of the proposed contextual augmentation approach. It achieves the effect of data augmentation by using a model to re-generate context for error patterns extracted from an existing parallel corpus.
    Wrong sentence and correct sentence are taken from the existing dataset.
    We extract error pattern of varying lengths from it.
    N-gram Aug represents the results of augmentation for error pattern of different lengths, 
    where source represents the ungrammatical sentence, target represents the corretion, red represents \textcolor{red}{grammatical errors} and green represents \textcolor{green!70!black}{correction results}.}
    \label{tab:case}
\end{table*}

There are two main reasons why previous synthetic data cannot apply to joint training in the fine-tuning phase.
(1) \textbf{Inconsistent Error Distribution}.
The high randomness of synthetic data makes it difficult to perfectly match the distribution of a certain high-quality data, leading joint training to performance degradation.
Rule-based substitution methods are limited by the distribution and word frequency of the unlabeled corpus.
Although model-based generation methods can generate different types of grammatical errors \cite{stahlberg2021synthetic}, they still do not have stable controllability for specific errors with a small amount of synthetic data.
As shown in Figure \ref{fig:distru}, the distribution of our proposed augmentation method is most consistent with the original dataset.
(2) \textbf{Noisy Label}.
Synthetic data is not human-labeled and cannot avoid introducing some mislabeling (inappropriate substitution or ungrammatical generation).
For example, "I think you are right" may be incorrectly annotated as "I think that you are right" in synthetic data.
As a text generation task with token-level metrics, the GEC task is very sensitive to this type of noise.
Directly joint training of synthetic and real data brings serious performance degradation \cite{zhang2019sequence}.
Recently, \citet{ye2023mixedit} propose the MixEdit framework for grammatical error augmentation of the fine-tuning stage through pattern replacement.
But it is still suffering from the two problems mentioned above.

In this paper, we propose a high-quality synthetic data construction method for the fine-tuning phase based on contextual augmentation.
It can be viewed as a combination of a rule-based substitution approach and a model-based generation approach, where the model is utilized to generate a rich context for the extracted error patterns.
An example of the augmentation data is shown in Table \ref{tab:case}.
Specifically, we first extract the error patterns (containing correct and incorrect token pairs) present in the real corpus through a GEC tool (ERRANT) and construct a corresponding error pattern pool.
After that, we sample error patterns from the pool based on the true frequency of the original dataset to regenerate contexts for them.
We regard this process as a hard constraint generation task, allowing the model to generate contextual sentences containing the correct pattern, and then obtaining the wrong sentence by rule substitution.
We attempt both GPT2 \cite{radford2019language} supervised generation and LLaMA2-7b-chat \cite{touvron2023llama} few-shot generation for our experiments.
Finally, we use the baseline GEC model to relabel the synthetic data for joint training to mitigate the noise in the synthetic data.



The main contributions of this paper can be summarized as follows:
\begin{itemize}
    \item We propose a synthetic data construction method based on contextual augmentation, which can stably generate a rich context for specific grammatical errors.
    \item To mitigate the effect of noisy labels,
    we introduce the re-labeling method into the synthetic data which improves the performance of the GEC model in joint training.
    \item Experiments show that our approach effectively enhances the robustness and performance of the GEC model by augmenting the high-quality annotated data in the fine-tuning phase.
\end{itemize}

We will release our code and model on github\footnote{\href{https://github.com/wyxstriker/CDA4GEC}{https://github.com/wyxstriker/CDA4GEC}}.


\begin{figure*}[ht]
    \centering
    \includegraphics[scale=0.55]{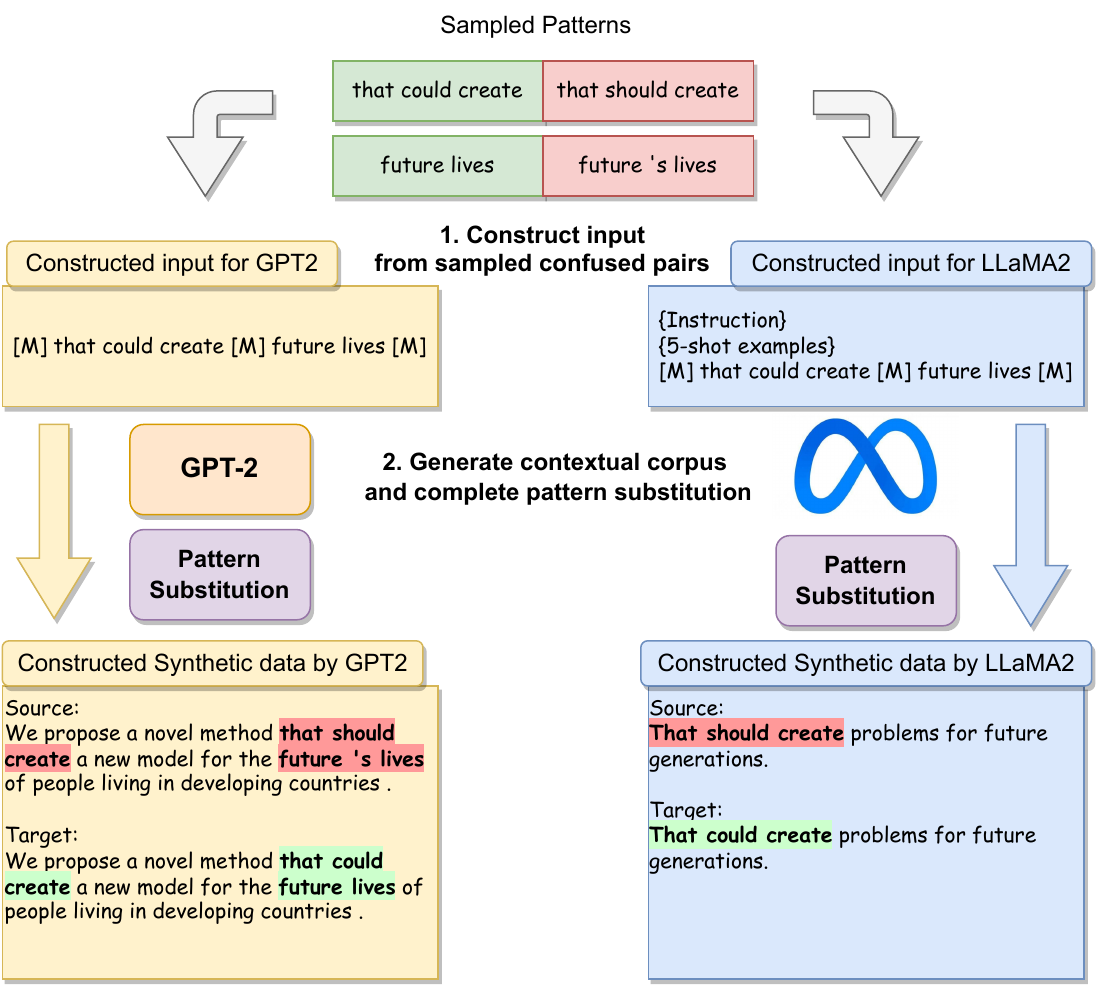}
    \caption{Illustration of synthetic data construction based on contextual augmentation. We uses both fine-tuned GPT2 and ICL of llama2 for the experiments. 
    The red in the sampling patterns represents the wrong pattern and the green represents the correct pattern.
    Note that we combine the sampled \textbf{correct} patterns into a certain format for context generation, followed by pattern substitution to obtain a parallel corpus. Due to the sample decoding strategy, there may be cases where the context does not fully cover the pattern in the input as in the case of LLaMA. In practice, we generate parallel corpus by directly ignoring the unmatched patterns.}
    \label{fig:flow}
\end{figure*}

\section{Method}

The main flow of our proposed synthetic data construction method based on context augmentation is illustrated in Figure \ref{fig:flow}.
First, we generate the synthetic data with contextual augmentation according to Section \ref{context_gen}'s method.
After denoising by relabeling (Section \ref{denoising}), we use the synthetic data to augment the original data in the joint training (Section \ref{trainstr}).

\subsection{Pattern-based Context Generation}
\label{context_gen}
Both rule-based substitution and model-based generation methods generate synthetic data that require large amounts of data (in the millions) to guarantee a wide range of errors \cite{kiyono2019empirical}.
However, in the supervised fine-tuning phase, the amount of data is very limited (W\&I+L only includes about 30k), and millions of synthetic data for joint training is unrealistic.
A more stable augmentation approach is needed to ensure that the original high-quality errors are adequately trained.

The main motivation for our proposed context augmentation is to leverage the modeling capability of language models to generate rich contexts for specific high-quality grammatical errors.
Compared with other synthesis methods, we ensure that the error distribution is consistent with the original dataset through rule-based error patterns and the diversity of sample contexts through model-based generation.
The synthesis method can be divided into three steps: building the error pattern pool (Section \ref{Patterns_extraction}), synthesizing the contextual corpus (Section \ref{context_gen_model}), and substituting \ref{pattern_sub} to get the parallel corpus.

\begin{figure}[t]
    \centering
    \includegraphics[scale=0.55]{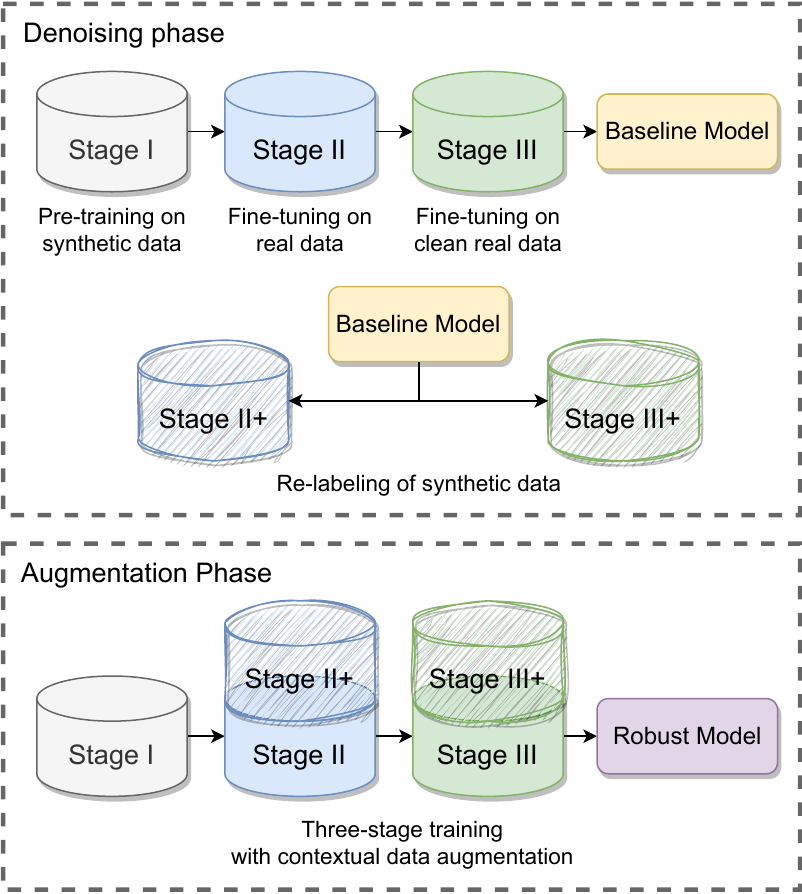}
    \caption{Illustration of the three phases of joint training with augmented data. We first denoise the synthetic data (Stage \uppercase\expandafter{\romannumeral2}+ \& \uppercase\expandafter{\romannumeral3}+) using a baseline model trained in three stages, and subsequently conduct joint training to obtain a robust model.}
    \label{fig:three_stage}
\end{figure}

\subsubsection{Error Patterns Extraction}
\label{Patterns_extraction}
We follow \citeposs{choe2019neural} setting and use the parsing tool ERRANT\footnote{https://github.com/chrisjbryant/errant} to extract the editing operations present in the parallel corpus as error patterns according to the rules.
We also extracted error patterns of different lengths for our experiments to ensure that the synthetic data contains more realistic errors.
Excessively long patterns will make it difficult to match them in unlabeled text with the original rule-based substitution method, but the contextual augmentation-based generation method can solve this problem well.

We finally extract the patterns for each human-labeled GEC dataset, and merge the corresponding patterns into an error pattern pool for the construction of synthetic data.
The statistics of extracted patterns can be found in Appendix \ref{pattern_num}.

\subsubsection{Contextual Corpus Generation}
\label{context_gen_model}
With the error pattern pool and the frequency of the corresponding errors, we can simply obtain a set of pattern datasets with the same distribution as the annotated corpus by sampling.
The goal of the generator model can be viewed as a hard-constrained text generation task \cite{welleck2019non}, generating a context that fully contains the target pattern.
Considering that existing pre-trained models are trained on grammatically correct corpora, we only generate the corresponding contextual corpus based on the correct patterns and subsequently construct the parallel corpus by rule-based substitution.

In particular, the input to the model will be a combination of several randomly sampled patterns, which can be formulated as:
\begin{equation} \label{gpt_input}
\label{eq_pattern}
Pattern_{input} = Pattern_1 \mbox{ } [M] \mbox{ } Pattern_2
\end{equation}
where $Pattern_i$ represents the correct pattern that was sampled and $[M]$ represents the context placeholder that needs to be generated.
It should be noted that the number of patterns for each sample will be randomly selected between 1 and 2, and Equation \ref{eq_pattern} represents the case of 2 patterns only.
The output of the model is a piece of text containing the corresponding input pattern.

In this paper, we experiment with two models as context generators, GPT2 and LLaMA2, representing the two settings of supervised fine-tuning and few-shot generation, respectively.

\paragraph{Finetuning for GPT2}
We choose GPT2 as the backbone network to represent the performance of the fine-tuned generative model in the contextual augmentation task.
The model generates the target corpus directly from the provided pattern, which can be formulated as:
\begin{equation} \label{gpt_input}
\label{eq_gpt_input}
S = Pattern_{input} \mbox{ <sep> } Sentence_{target}
\end{equation}
where $Pattern_{input}$ is the combination of the patterns mentioned in Equation \ref{eq_pattern} and $Sentence_{target}$ is the target corpus containing all the patterns.
$\mbox{<sep>}$ is the special token dividing the input $S$ into two parts.

For the training phase, we use the autoregressive way consistent with the pre-training:
\begin{equation} \label{gpt_input}
\label{eq_gpt_loss}
\mathcal{L} = 
\sum_{k=i}^{j} -log(P(t_k|t_0t_1...t_{k-1};\theta))
\end{equation}
where $\theta$ is the set of parameters of the language model, $i$ and $j$ represent the start and the end index of $Sentence_{target}$, and $t_i$ represents the i-th token in the model input $S$ like Equation \ref{eq_gpt_input}.

As for the training data, in order to ensure that the style of the generated text is consistent with the training data, we directly adopt the correct sentences in the non-native speaker GEC dataset C-Lang8 \cite{rothe2021simple} dataset as the target sentences to construct the training set.
Specifically, we randomly replace multiple consecutive text segments in the corpus with $[M]$ label and train the model to generate the corresponding context based on the remaining text segments (error patterns during inference).

\paragraph{Few-shot generating for LLaMA2}
Recently, LLMs \cite{brown2020language, wei2021finetuned, touvron2023llama} have presented powerful in-context learning capabilities to accomplish complex NLP tasks based on a few example samples.
Therefore we also try to use the LLM to generate a more appropriate context for the error patterns.
We directly use the prompt with the 5-shot setting and ask the LLM to generate the conditional corpus.
Details of the prompts and input format can be found in Appendix \ref{prompt}.
In addition, the GEC corpus is usually large, and considering the cost issue, we choose the open-source LLM LLaMA2-7b-chat \cite{touvron2023llama} for our experiments.

        

\subsubsection{Pattern Substitution}
\label{pattern_sub}
Given the contextual corpus and error pattern pairs, we can obtain the corresponding GEC parallel corpus by simple substitution \cite{choe2019neural}.
To ensure the diversity of the corpus, we choose the sampling decoding strategy during generation, and we directly ignore the patterns that can't be matched exactly.
Besides, we only substitute the error patterns for 50\% of the synthetic data to ensure a consistent error rate with the annotated datasets during the joint training process.


%


\subsection{Synthetic Data Denoising}
\label{denoising}
Compared to human-annotated data, synthetic data is more accessible but inevitably noisy.
Previous work \cite{zhang2019sequence} has proven that direct joint training of synthetic and real data affects the metrics of the final model.
Improper substitutions (grammatically correct both before and after substitution) are the main cause of noisy synthetic data.
\citet{yasunaga2021lm, cao2023unsupervised} propose some sentence-level filtering methods based on scores such as PPL, but the filtering granularity and accuracy are not sufficient in the joint training setting.
We need an efficient way of filtering at the token level.

Inspired by \citeposs{rothe2021simple,ye2023system} distillation method, which mitigates the effects of noisy data by relabeling the corpus with a powerful GEC model, we also want to denoise the corpus through relabeling.
Specifically, we view the synthetic data as an unlabeled grammatical error-filled corpus and relabel it using a strong baseline model.
Since the synthetic data is obtained by augmentation using the original dataset, relabeling using the original model effectively removes the noise while correcting most of the grammatical errors.

\subsection{Joint Training Process}
\label{trainstr}
To obtain a strong baseline model, we follow \citeposs{bout2023efficient} approach of using three stages for training.
Our proposed data augmentation method will also be applied to the stages of fine-tuning.

As shown in Figure \ref{fig:three_stage}, we divide the available dataset into three stages for training.
We use C4$_{200M}$ \cite{stahlberg2021synthetic} dataset for the pre-training phase (stage \uppercase\expandafter{\romannumeral1}),
which generate grammatical errors with type distribution consistent with BEA-Dev \cite{bryant2019bea} based on the seq2edit model.
For Stage \uppercase\expandafter{\romannumeral2}, we used the complete available annotated dataset (see Table \ref{tab:dataset} for details) to fine-tune the model, including Lang-8, NUCLE, FCE, and W\&I+L.
As the highest-quality annotated GEC dataset, we individually fine-tune the stage \uppercase\expandafter{\romannumeral3} on W\&I+L.

Due to data distribution and quality, previously synthetic data are mainly utilized in the pre-training phase (stage \uppercase\expandafter{\romannumeral1}).
In contrast, our proposed context augmentation approach is mainly used to adapt a small amount of high-quality fine-tuned data (stage \uppercase\expandafter{\romannumeral2}\&\uppercase\expandafter{\romannumeral3}).
We generate different amounts of synthetic data (Stage\uppercase\expandafter{\romannumeral2}-syn and Stage\uppercase\expandafter{\romannumeral3}-syn in Table \ref{tab:dataset}) for different stages using the corresponding error pattern pool.
After that, we directly train the synthesized data jointly with the real data of fine-tuning stages, as shown in Figure \ref{fig:three_stage}.

\begin{table}[t]
    \centering
    \scalebox{0.9}{
    \begin{tabular}{lccc}
    \toprule
    Dateset & Errorful\% & Sentences\# & Usage \\
    \midrule
    C4$_{200M}^{\star}$   & 99.4 & $\sim$180M  & \uppercase\expandafter{\romannumeral1} \\
    Lang-8    & 48.0 & 1,037,561 & \uppercase\expandafter{\romannumeral2}\\
    NUCLE    & 38.0 &  56,958 & \uppercase\expandafter{\romannumeral2}\\
    FCE    & 62.5 & 28,350 & \uppercase\expandafter{\romannumeral2}\\
    W\&I+L    & 67.3 & 34,304 & \uppercase\expandafter{\romannumeral2}\&\uppercase\expandafter{\romannumeral3}\\
    \midrule
    Stage\uppercase\expandafter{\romannumeral2}-Syn$^{\star}$ & 50.0 & 2M & \uppercase\expandafter{\romannumeral2}+\\
    Stage\uppercase\expandafter{\romannumeral3}-Syn$^{\star}$ & 50.0 & 200,000 & \uppercase\expandafter{\romannumeral3}+\\
    \midrule
    BEA19-Dev & 64.3 & 4,384 & Dev\\
    Conll14-Test & 71.9 & 1,312 & Test\\
    BEA19-Test & N/A & 4,477 & Test\\
    \bottomrule
    \end{tabular}
}
    \caption{Statistical information on grammatical error correction datasets. Note that $\star$ indicates synthetic datasets. \uppercase\expandafter{\romannumeral2}+ and \uppercase\expandafter{\romannumeral3}+ represent the augmented dataset of corresponding stages, which will be mixed with real data for joint training in the proposed method.}
    \label{tab:dataset}
\end{table}

\begin{table*}[ht]
\centering
\scalebox{0.9}{
\begin{tabular}{lccccccc}
\toprule
\multirow{2}{*}{\textbf{Model}} & \multirow{2}{*}{\textbf{Model Size}} & \multicolumn{3}{c}{\textbf{CoNLL2014}} & \multicolumn{3}{c}{\textbf{BEA19-Test}} \\
 & & $Prec$ & $Rec$ & $F_{0.5}$ & $Prec$ & $Rec$ & $F_{0.5}$\\
\midrule
GECToR$^\diamondsuit$ \cite{omelianchuk2020gector} & 350M & 77.5 & 40.1 & 65.3 & 79.2 & 53.9 & 72.4 \\
T5-large$^\heartsuit$ \cite{rothe2021simple} & 770M & - & - & 66.1 & - & - & 72.1 \\
T5-XL$^\heartsuit$ \cite{rothe2021simple} & 3B & - & - & 67.8 & - & - & 73.9 \\
T5-XXL$^\heartsuit$ \cite{rothe2021simple} & 11B & - & - & \textbf{68.9} & - & - & \textbf{75.9} \\
ShallowAD$^\clubsuit$ \cite{sun2021instantaneous} & $\sim$240M & 71.0 & 52.8 & 66.4 & - & - & 72.9\\
SynGEC$^{\heartsuit}$ \cite{zhang2022syngec} & 400M & 74.7 & 49.0 & 67.6 & 75.1 & 65.5 & 72.9\\
TemplateGEC$^\heartsuit$ \cite{li2023templategec} & 770M & 74.8 & 50.0 & 68.1 & 76.8 & 64.8 & 74.1 \\
MixEdit$^\heartsuit$ \cite{ye2023mixedit} & 400M & 75.6 & 46.8 & 67.3 & 76.4 & 62.7 & 73.2\\
MultiTaskBART$^\spadesuit$ \cite{bout2023efficient} & 400M & 75.4 & 51.2 & \textbf{68.9} & 78.2 & 65.5 & 75.3\\
\midrule
BART Baseline$^\spadesuit$ & 400M & 73.8 & 53.5 & 68.6 & 74.5 & 68.9 & 73.5\\
+ CDA w/o denoising  & 400M & 76.7 & 44.8 & 67.1 & 76.1 & 67.3 & 74.1 \\
+ CDA w/ denoising  & 400M & 76.2 & 52.2 & \textbf{69.8} & 77.7 & 67.5 & \textbf{75.4} \\
\bottomrule
\end{tabular}
}
\begin{tabular}{lc}
\hline
\end{tabular}
\caption{The results of the strong BART Baseline initialized on C4$_{200M}$ and Context Date Augmentation (CDA) methods for the single model.
CDA w/o denising means directly using the raw synthetic data constructed by the proposed method without relabeling.
In addition to publicly available annotated datasets, existing GEC models also use:
$\diamondsuit$ rule-based synthetic data from one-billion-word (9M),
$\heartsuit$ cleaned version of Lang8 (2.4M),
$\clubsuit$ model-based synthetic data (300M),
$\spadesuit$ model-based synthetic data (C4$_{200M}$).
}
\label{tab:main_res}
\end{table*}

\begin{figure*}[h]
    \centering
    \includegraphics[scale=0.5]{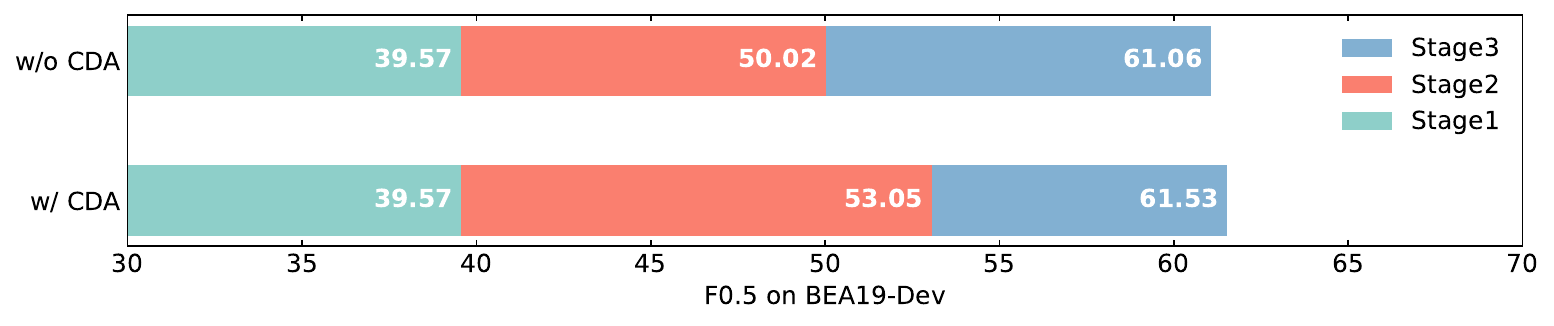}
    \caption{The results of the three-stage model on BEA19-Dev after contextual augmentation respectively.}
    \label{fig:hex}
\end{figure*}

\section{Experiment}
\subsection{Setting}
\paragraph{Datasets}
In Table \ref{tab:dataset}, we summarize the statistical information for the all relevant datasets.
The C4$_{200M}$ \cite{stahlberg2021synthetic} dataset used for stage \uppercase\expandafter{\romannumeral1} is synthetic data based on Seq2Edit \cite{stahlberg2020seq2edits} model generation.
For the other stages, we use the following common GEC datasets:
Lang-8 Corpus of Learner English (Lang-8) \cite{mizumoto2011mining, tajiri2012tense} collects from non-native speaker online learning websites Lang-8\footnote{https://lang-8.com/};
National University of Singapore Corpus of Learner English (NUCLE) \cite{dahlmeier2013building} consists of essays written by undergraduate students on a variety of topics and annotated by professional English teachers;
First Certificate in English (FCE) \cite{yannakoudakis2011new} primarily contains answers to the upper-intermediate level exams written by English language learners;
Write \& Improve + LOCNESS Corpus (W\&I+L) \cite{bryant2019bea} includes two parts of data.
The Write \& Improve dataset consists of chunks of text (articles, letters, etc.) submitted to the W\&I system written by English learner;
In contrast, LOCNESS consists of essays written by native English-speaking students and is used for evaluation purposes only.

In addition to the existing publicly available datasets, we constructed synthetic data for different stages (corresponding to StageX-Syn in the Table \ref{tab:dataset}) by contextual augmentation using the proposed method.
As for the amount of synthetic data, we heuristically choose 2M pairs for Stage \uppercase\expandafter{\romannumeral2}, 200,000 pairs for Stage \uppercase\expandafter{\romannumeral3}.
We perform ablation experiments on the amount of augmented data in subsequent analyses.

\paragraph{Evaluation}
We use BEA19-Dev \cite{bryant2019bea} as a validation set to evaluate the performance of the GEC model.
In the main experiments, we report results of Conll14-Test \cite{ng2014conll} using the official M2 scorer \cite{dahlmeier2012better}, and results of BEA19-Test \cite{bryant2019bea} using ERRANT \cite{bryant2017automatic} on the online platform\footnote{https://codalab.lisn.upsaclay.fr/competitions/4057}.

\paragraph{Training Details}
As described in Section \ref{context_gen_model}, we use GPT2-base and LLaMA2-7b-chat as context generators for our experiments.
For the implementation of the baseline GEC model, we refer to previous setups \cite{zhang2022syngec} and use Seq2Seq-based BART-large \cite{lewis2019bart} for the experiments, which is trained using the Fairseq\footnote{https://github.com/facebookresearch/fairseq} framework.
More details on hyperparameters can be found in Appendix \ref{hyperparameters}.

\subsection{Baseline Approaches}

We select several recent state-of-the-art methods as baselines for comparison.
GECToR \cite{omelianchuk2020gector} is an efficient auto-encoder grammatical error correction model, which corrects errors by predicting edit tags. \citet{rothe2021simple} verify the performance of T5 \cite{raffel2020exploring} models of various scales (from small to xxl) on the GEC task.
SynGEC \cite{zhang2022syngec} incorporates the syntactic information of the text into the model using GCN.
TemplateGEC \cite{li2023templategec} fuses the seq2edit and seq2seq models to provide a new two-stage framework for error detection and correction.
\citet{ye2023mixedit} propose a data augmentation approach MixEdit that strategically and dynamically augments realistic data, without requiring extra monolingual corpora.
\citet{bout2023efficient} propose a multi-task pre-training method and optimization strategy, which greatly improved the performance of the GEC model.

In this paper, we use the three-stage training model initialized by the C4$_{200M}$ datasets as strong baselines.
With Contextual Data Augmentation (CDA), we integrate contextually augmented synthetic data for training in the fine-tuning phase, as shown in Figure \ref{fig:three_stage}.

\subsection{Main Experimental Results}

The experimental results of our proposed augmentation method on CoNLL14 and BEA19 are shown in Table \ref{tab:main_res}.
We obtain a strong baseline model by training in three stages according to the \citeposs{bout2023efficient} setting.
The results show that contextual data augmentation can effectively improve the robustness and generalization of the original model, and bring significant improvements on both CoNLL14 and BEA19-Test datasets.
Our 400M BART model achieves the state-of-the-art through contextual data augmentation, and is comparable to the 11B T5-XXL.
In addition to this, we find that the impact of the augmented data's noisy labels can be well mitigated by simple relabeling.
It should be noted that the proposed method improves the modeling precision with a slight loss in recall, which is encouraged in GEC tasks since ignoring an error is not as bad as proposing a wrong correction \cite{ng2014conll}.

In addition to this, we have analysed the impact of the proposed data augmentation approach on the model at different stages.
As shown in Table \ref{fig:hex}, the enhancement of model effectiveness by contextual data augmentation is more pronounced in the second stage where the quality of annotated data is relatively low.

\section{Analysis}


\subsection{Impact of Different Generators}
\label{model_scale}

In this article, we have experiment with two generator settings, GPT2 fine-tuning, and LLaMA2 ICL, to generate synthetic data.
The GPT2 fine-tuning model is relatively small, which has a faster generation efficiency and follows the task requirements better after training.
On the contrary, LLaMA2 has a larger number of parameters and generates more diverse and fluent texts. But the model generates more slowly and follows the instructions more weakly.

To verify the effect of the different generators on the quality of the generated text, we use them to generate 200k synthetic data on the high-quality text of stage \uppercase\expandafter{\romannumeral3} respectively for joint training.
The experiment results are shown in Table \ref{tab:llama}.
We find similar conclusions to \citet{xu2023s2ynre}, that whether the synthetic data generated by the fine-tuned model or the LLM in-context learning has little effect on the final model performance.
So we mainly use GPT2 fine-tuning as the generator for experiments for efficiency considerations.
\begin{table}[t]
    \centering
    \scalebox{0.9}{
    \begin{tabular}{lccc}
    \toprule
    \multirow{2}{*}{\textbf{Method}}   & \multicolumn{3}{c}{\textbf{BEA19-Dev}} \\
     & $Prec$ & $Rec$ & $F_{0.5}$ \\
    \midrule
    Stage2 Model & 61.28 & 28.84 & 50.02 \\
    + Stage3 GPT2 & 64.16 & \textbf{51.13} & 61.05 \\
    + Stage3 LLaMA2 & \textbf{64.23 }& 51.07 & \textbf{61.08} \\
    \midrule
    Stage2 Model & 61.28 & 28.84 & 50.02 \\
    + Stage3 1-gram & 63.73 & \textbf{52.50} & 61.12 \\
    + Stage3 3-gram & \textbf{64.39} & 51.07 & \textbf{61.20} \\
    + Stage3 5-gram & 64.16 & 51.13 & 61.05 \\
    \bottomrule
    \end{tabular}}
    \caption{Ablation study of the different generators and the different pattern lengths on BEA19-Dev.}
    \label{tab:llama}
\end{table}


\label{data_amount}

\subsection{Impact of Pattern Length}
\label{pattern_len}
Unlike previous rule-based substitution approaches \cite{choe2019neural}, our proposed method is not restricted to the distribution of the unlabeled corpus, so the length of the error pattern is no longer limited to the token level.
To obtain the optimal pattern length, we experiment with contextual augmentation using 1-gram, 3-gram, and 5-gram patterns as shown in Table \ref{tab:case}.
The results are shown in Table \ref{tab:llama}, the model has the best performance with synthetic data generated by the 3-gram pattern.
We hypothesize that appropriately long patterns make the synthetic data distribution closer to the source data, indirectly improving the quality of the relabeling.

\begin{figure}[t]
    \centering
    \includegraphics[scale=0.5]{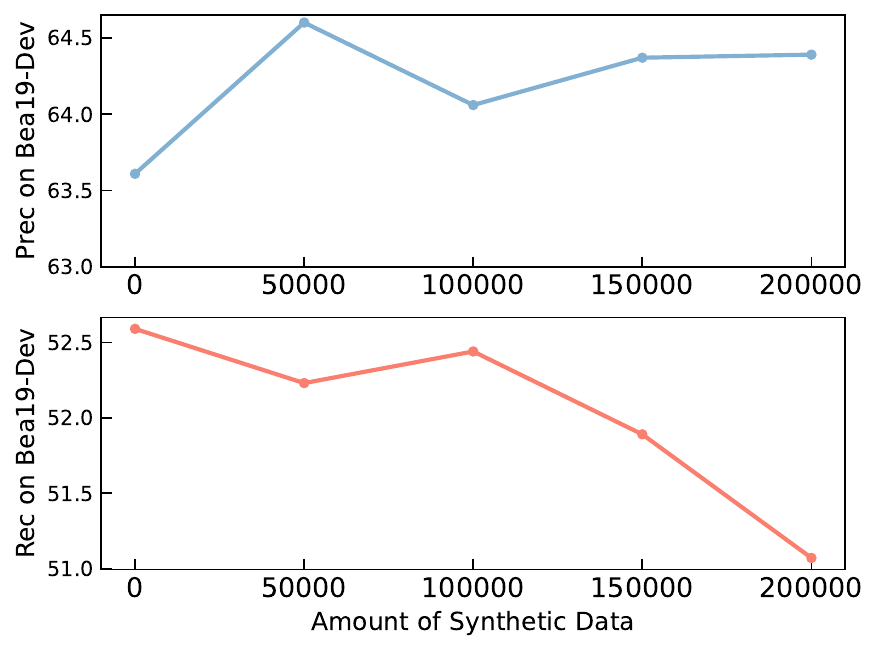}
    \caption{The effect of different amounts of synthetic data in joint training on the final system.}
    \label{fig:rate}
\end{figure}

\subsection{Impact of Data Mixing Ratio}
In joint training, the impact of the ratio of synthetic to real data is significant.
We conduct joint training experiments with different amounts of synthetic data, and the results are shown in Figure \ref{fig:rate}.
Consistent with our analysis, the ratio of synthetic data to real data is a trade-off.
As the amount of synthetic data increases, the precision of the system gradually increases.
At the same time, the proportion of higher-quality real data has declined, possibly leading to some decline in error recall.

\subsection{Analysis of Error Categories}
To evaluate the specific advantages of the proposed data augmentation approach, we have conducted experiments on different error categories of error correction.
As shown in Table \ref{tab:type}, the CDA method provides a richer context for low-frequency errors, allowing the model to achieve great improvements in this type of error. In contrast, high-frequency errors themselves have a large amount of real training data, and the addition of synthetic data may have a slight impact on the original results such as orthographic (ORTH) and punctuation (PUNCT). This conclusion is consistent with \citet{rothe2021simple}, indicating that the distillation model does not handle these two types of errors well.

\begin{table}[t]
    \centering
    \tabcolsep=0.45cm
    \scalebox{0.9}{
    \begin{tabular}{lcc}
    \toprule
    \textbf{Error Type} & \textbf{Baseline} & \textbf{+CDA} \\
    \midrule
    PUNC & \textbf{77.26} & 76.96 \\
    DET & 79.04 & \textbf{80.42} \\
    PREP & 74.92 & \textbf{76.84} \\
    OTHER & \textbf{84.23} & 82.32 \\
    SPELL & 89.61 & \textbf{91.51} \\
    \midrule
    CONTR & 91.67 & \textbf{93.02} \\
    ADJ & 58.44 & \textbf{62.15} \\
    ADV & 59.43 & \textbf{62.72} \\
    CONJ & 48.39 & \textbf{53.33} \\
    NOUN & 45.39 & \textbf{48.57} \\
    \bottomrule
    \end{tabular}}
    \caption{Experimental results on different grammatical error categories. We show the $F_{0.5}$ results for the highest-frequency five error categories (top) and the lowest-frequency five errors categories (bottom).
    The error categories refer to the labelling of the ERRANT tool.}
    \label{tab:type}
\end{table}

\subsection{Quality Assessment of Synthetic Data}
To compare the quality of our proposed context augmentation with other synthetic data,
we adopt the same three-stage training and denoising for synthetic data of different construction methods.
We choose PIE \cite{awasthi2019parallel} and C4$_{200M}$ \cite{stahlberg2021synthetic} as our baselines, which represent the rule-based substitution method and the model-based generation method, respectively.
The results are shown in Table \ref{tab:syn}, where our proposed CDA method is able to better fit the distribution of the original dataset within the limited data, thus improving the model performance in joint training.


\begin{table}[t]
    \centering
    \scalebox{0.9}{
    \begin{tabular}{lccc}
    \toprule
     \multirow{2}{*}{\textbf{Synthetic Data}}   & \multicolumn{3}{c}{\textbf{BEA19-Dev}} \\
     & $Prec$ & $Rec$ & $F_{0.5}$ \\
    \midrule
    N/A & 63.61 & 52.59 & 61.06 \\
    \midrule
    PIE & 63.89 & 51.68 & 61.01 \\
    C4$_{200M}$ & 63.96 & 51.42 & 60.98 \\
    CDA (ours) & \textbf{64.39} & \textbf{52.23} & \textbf{61.53} \\
    \bottomrule
    \end{tabular}}
    \caption{Performance of different synthetic data participating in joint training after denoising.
    N/A means the no data augmentation situation.}
    \label{tab:syn}
\end{table}

\section{Related Work}
\subsection{Synthetic Data for GEC Task}
\paragraph{Construction of Synthetic Data}
As a data-starved field, synthetic data has been proven effective in improving the GEC systems \cite{kiyono2019empirical},
which can be categorized into rule-based substitution and model-based generation methods.
Rule-based methods are mainly constructed through direct noise addition \cite{xu2019erroneous,zhou2019improving,kiyono2020massive}, pattern substitution \cite{choe2019neural}, and parsing tools \cite{grundkiewicz2019neural}.
Model-based generation methods mainly include back-translation \cite{xie2018noising,stahlberg2021synthetic} and round-trip translation \cite{zhou2019improving} based on Seq2Seq architecture.
\cite{stahlberg2021synthetic} propose a synthesis method based on the Seq2Edit \cite{stahlberg2020seq2edits} architecture capable of generating synthetic data by specifying grammatical error types.

\paragraph{Utilization of Synthetic Data}
\citet{zhang2019sequence} have explored the use of synthetic data through detailed experiments. The experiments prove that using synthetic data in the pre-training phase achieves optimal results.
Some unsupervised grammatical error correction work \cite{yasunaga2021lm, cao2023unsupervised} have used the self-training framework to label and co-train unlabeled error corpus to obtain an improvement in effectiveness.

\subsection{LLM for GEC Task}
LLMs \cite{brown2020language, wei2021finetuned, touvron2023llama} have made significant improvements in a wide range of natural language processing tasks.
However, LLMs do not perform well on common benchmarks \cite{coyne2023analyzing} due to traditional evaluation metrics and over-corrections.
Although \citet{fang2023chatgpt} have demonstrated that some improvement is achieved by few-shot and chain-of-thought settings, there is still a big gap between LLMs and traditional fine-tuning models.
In view of this, using LLM to construct synthetic data for the GEC task \cite{fan2023grammargpt} can be considered as another feasible direction.

\section{Conclusion}
In this paper, we propose a synthetic data construction method based on contextual augmentation.
It stably augments the context of the source data and ensures a consistent error distribution.
Previous methods suffer from noisy labels of synthetic data.
We significantly improve the performance of synthetic data in joint training through a re-labeling-based denoising method.
We validate the effectiveness of our proposed method on several common datasets.

\section*{Limitations}
Firstly, compared to other current synthetic data construction methods, generating synthetic data based on contextual augmentation takes more time and resources. For each sample, we need to complete the inference process on both sides of context generation and denoising.
Secondly, the re-predicted results are not completely correct and can only alleviate noise. There are still a small number of incorrect labels in the synthetic data.
In addition, it should be noted that the context augmentation method we proposed can only provide richer context for errors existing in the annotated dataset, and cannot introduce new grammatical errors.
We will focus on investigating how to better generate high-quality synthetic data that contains a wider variety of grammatical errors utilizing LLMs in our future work.

\section*{Ethics Statement}

In this paper, we explore the application of contextual augmentation-based synthetic data on the GEC task.
The source data for these methods come exclusively from publicly available project resources on legitimate websites and do not involve any sensitive information. 
In addition, all baselines and datasets used in our experiments are also publicly available, and we have acknowledged the corresponding authors by citing their work.

\section*{Acknowledgements}

We gratefully acknowledge the support of the National Natural Science Foundation of China (NSFC) via grant 62236004 and 62206078.

\bibliography{acl_latex}

\clearpage
\appendix

\section{Error Pattern Information}
\label{pattern_num}
We have extracted the error patterns from the existing annotated dataset using the ERRANT tool \cite{bryant2017automatic}, as described in Section \ref{Patterns_extraction}.
The number of patterns for each dataset is shown in Table \ref{tab:pattern_pool}.
We maintain a separate error pattern pool for each dataset, from which we sample each time to generate synthetic data.

\begin{table}[h]
    \centering
    \begin{tabular}{ccc}
    \toprule
     Dataset & Sentence\# & Error Pattern\# \\
    \midrule
    Lang-8    & 1,037,561 & 677,475\\
    NUCLE    & 56,958 & 49,347\\
    FCE    & 28,350 & 43,854\\
    W\&I+L    & 34,304 &  62,952\\
    \bottomrule
    \end{tabular}
    \caption{Statistics of the error pattern pool for each dataset.}
    \label{tab:pattern_pool}
\end{table}

\begin{table}[ht]
    \centering
    \begin{tabular}{lc}
    \toprule
    Configuration & Value \\
    \midrule
    \multicolumn{2}{c}{\textbf{Stage1}} \\
    \midrule
    Backbone  &  BART-large \\
    & \cite{lewis2019bart} \\
    Devices & 4 Tesla V100S-PCIE-32GB \\
    Epochs & 10 \\
    Max tokens & 4096\\
    Update freq & 8 \\
    Optimizer & Adam \\
    & \cite{kingma2014adam}\\
    Learning rate & 3e-05\\
    Max source length & 1024 \\
    Dropout-src & 0.2 \\
    Clip norm & 0.1 \\
    Label smoothing & 0.1 \\
    \midrule
    \multicolumn{2}{c}{\textbf{Stage2}} \\
    \midrule
    Epochs & 20 \\
    Learning rate & 1e-05\\
    Warmup-updates & 2000 \\
    Patient & 5 \\
    \midrule
    \multicolumn{2}{c}{\textbf{Stage3}} \\
    \midrule
    Epochs & 50 \\
    Learning rate & 3e-06\\
    Warmup-updates & 200 \\
    Patient & 10 \\
    \bottomrule
    \end{tabular}
    \caption{Hyperparametric details of the BART-based three-stage GEC model. In Stage \uppercase\expandafter{\romannumeral2},\uppercase\expandafter{\romannumeral3} only the parameters that differ from those in Stage1 are described.}
    \label{tab:hp_bart}
\end{table}

\section{Hyper-parameters}
\label{hyperparameters}
We illustrate the hyper-parameters during training of the baseline model (see Table \ref{tab:hp_bart} for details) and the GPT2-based generative model (see Table \ref{tab:hp_gpt} for details) here.

\begin{table}[h]
    \centering
    \begin{tabular}{lc}
    \toprule
    Configuration & Value \\
    \midrule
    Backbone  &  GPT2-base \\
    & \cite{radford2019language} \\
    Devices & 4 Tesla V100S-PCIE-32GB \\
    Epochs & 20 \\
    Batch size & 32\\
    Update freq & 4 \\
    Optimizer & AdamW \\
    & \cite{loshchilov2017decoupled}\\
    Learning rate & 5e-05\\
    Max length & 256 \\
    Warmup ratio & 0.1 \\
    \bottomrule
    \end{tabular}
    \caption{Hyperparametric details of the GPT2-based contextual generator.}
    \label{tab:hp_gpt}
\end{table}

\section{Instruction Format for Synthetic Data Generation}
\label{prompt}
When using LLaMA2-7b-chat for synthetic data generation, we use the 5-shot setting to generate the corresponding context for the sampled error patterns.
We have given an example to illustrate the specific input format as shown in Table \ref{tab:prompt_case}.

\begin{table*}[h]
    \centering
    \begin{tabularx}{0.95\textwidth}{c|X}
    \hline
    Instruction & [INST] < <SYS> > You are a helpful assistant.< </SYS> > \\
    & Use phrases from \#input to make sentences. \\
    & You should fill in [M] to make \#input sentence more complete. \\
    & You can't change any form or order of the words in \#input. \\
    & Make sure you fully use the phrases in \#input. [/INST] \\
    \hline
    5-shot &  \#input: [M] sized city with eighty thousand [M] \\
    & \#output: My town is a medium - sized city with eighty thousand inhabitants . \\
    & \\
    & \#input: [M] my own plan too , [M] to be the same as them . [M] \\
    & \#output: I have my own plan too , but I do n't want to be the same as them . I want to become a journalist . \\
    & \\
    & \#input: Nowadays , each family has more than 1 [M] one of several reasons why [M] \\
    & \#output: Nowadays , each family has more than 1 car for each person , this is only one of several reasons why people use less public transport . \\
    & \\
    & \#input: [M] they might want to safeguard [M] \\
    & \#output: On the other hand , they might want to safeguard the national image . \\
    & \\
    & \#input: Lucy , Molly , and [M] a cowboy , and a [M] \\
    & \#output: Lucy , Molly , and their parents , a cowboy , and a teacher . \\
    \hline
    Input & \#input: And I went [M] important [M]\\
    \hline
    Output & \#output: And I went to the library to study for an important exam .\\
    \hline
    \end{tabularx}
    \caption{An example input format for LLaMA2 ICL synthetic data generation}
    \label{tab:prompt_case}
\end{table*}

\end{document}